\documentclass[times,twocolumn,final]{elsarticle}
\usepackage[ruled,vlined]{algorithm2e}
\SetKwInput{KwInput}{Input}
\SetKwInput{KwRequire}{Require}
\SetKwInput{KwOutput}{Output}

\usepackage{medima}
\usepackage{framed,multirow}
\usepackage{graphicx}

\usepackage{amssymb}
\usepackage{latexsym}

\usepackage{url}
\usepackage{xcolor}
\usepackage{float} 
\usepackage{caption}
\usepackage{arydshln}
\makeatletter

\newcommand{\Rmnum}[1]{\expandafter\@slowromancap\romannumeral #1@}
\makeatother

\usepackage{tikz}
\usepackage[T1]{fontenc}
\usepackage{hyperref}
\usepackage{amsmath} 
\usepackage{booktabs}
\usepackage{multirow}
\usepackage{threeparttable}
\definecolor{newcolor}{rgb}{.8,.349,.1}

\journal{Medical Image Analysis}

\begin{document}

\verso{Y.Yao \textit{et~al.}}

\begin{frontmatter}

\title{From Uncertainty to Clarity: Uncertainty-Guided Class-Incremental Learning for Limited Biomedical Samples via Semantic Expansion}%

\author[1]{Yifei \snm{Yao}\fnref{fn1}}
\ead{yifei3.23@intl.zju.edu.cn}
\fntext[fn1]{Equal contribution.}
\author[2]{Hanrong \snm{Zhang}\fnref{fn1}}
\ead{hanrong.22@intl.zju.edu.cn}
\author[2]{\snm{}}
\author[1]{\snm{}\corref{cor1}}
\cortext[cor1]{Corresponding author:}

\address[1]{Centre of Biomedical Systems and Informatics of Zhejiang University-University of Edinburgh Institute (ZJU-UoE Institute), International Campus, Zhejiang University, Haining, 314400, Zhejiang, China}
\address[2]{Zhejiang University and the University of Illinois Urbana–Champaign Joint Institute, Haining, 314400, Zhejiang, China}


\begin{abstract}
In real-world clinical settings, data distributions evolve over time, with a continuous influx of new, limited disease cases. Therefore, class incremental learning is of great significance, i.e., deep learning models are required to learn new class knowledge while maintaining accurate recognition of previous diseases. However, traditional deep neural networks often suffer from severe forgetting of prior knowledge when adapting to new data unless trained from scratch, which undesirably costs much time and computational burden. Additionally, the sample sizes for different diseases can be highly imbalanced, with newly emerging diseases typically having much fewer instances, consequently causing the classification bias. To tackle these challenges, we are the first to propose a class-incremental learning method under limited samples in the biomedical field. First, we propose a novel cumulative entropy prediction module to measure the uncertainty of the samples, of which the most uncertain samples are stored in a memory bank as exemplars for the model's later review. Furthermore, we theoretically demonstrate its effectiveness in measuring uncertainty. Second, we developed a fine-grained semantic expansion module through various augmentations, leading to more compact distributions within the feature space and creating sufficient room for generalization to new classes. Besides, a cosine classifier is utilized to mitigate classification bias caused by imbalanced datasets. Across four imbalanced data distributions over two datasets, our method achieves optimal performance, surpassing state-of-the-art methods by as much as 53.54\% in accuracy.
\end{abstract}


\end{frontmatter}


\section{Introduction}
\label{introduction}
Recent years have witnessed rapid and extraordinary progress in the development of machine learning~\citep{vats2024incremental, SAKA}. This progress has played a critical role in various tasks of biomedical image analysis by effectively capturing the regularities or patterns inherent in data, relying on vast amounts of annotated data and continuously advancing models~\citep{shen2017deep, brody2013medical}. Whereas in a real-world clinical setting, data distribution gradually changes over time, new virus strains may emerge, and new symptoms may need to be identified.

\begin{figure}[!t]
\centering
\includegraphics[width=0.5\textwidth]{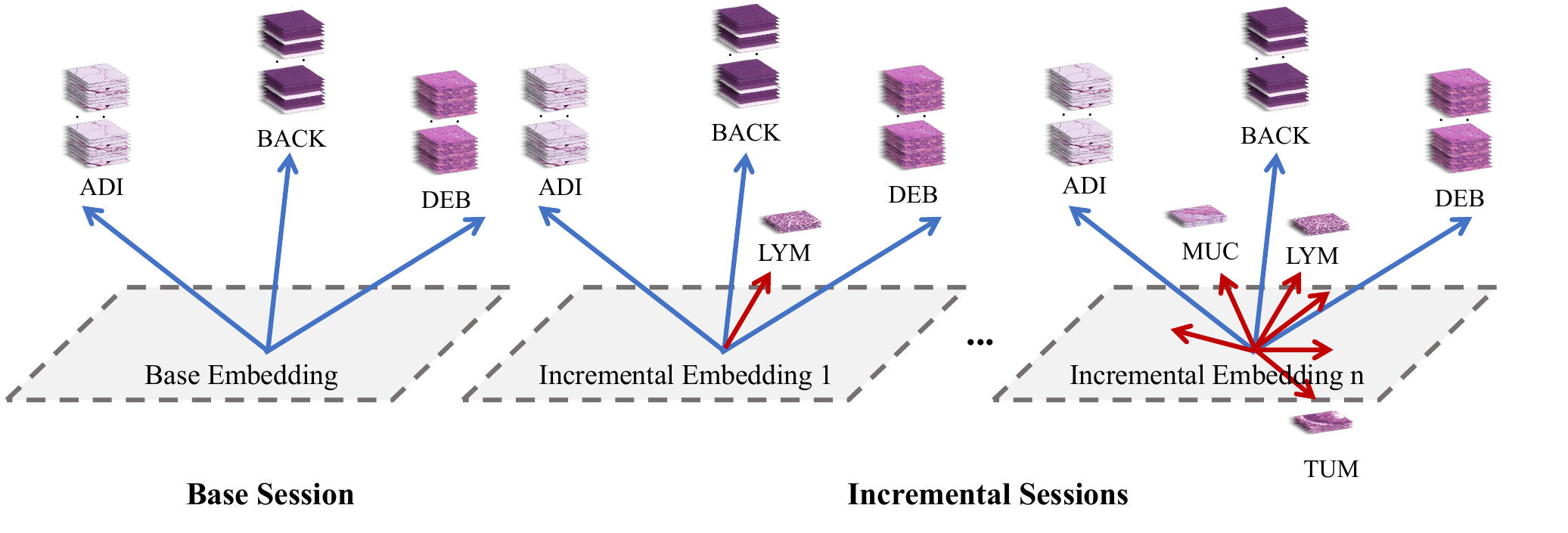}
\caption{A class-incremental learning process, where a classification model initially trained on a large set of images from three tissue types—ADI (adipose tissue), BACK (background), and DEB (debris)—is gradually updated by incorporating a small number of images from other types, such as LYM (lymphocytes), with corresponding updates to the model's embedding space.}
\label{fig:intro}
\end{figure}

For the natural visual systems of humans and animals, they can continuously absorb new visual information while simultaneously preserving previously acquired knowledge~\citep{gao2023incremental}. In the realm of machine learning, models also strive for a similar capability known as continuous learning, also referred to as continual learning or lifelong learning, by gradually updating themselves with new cases~\citep{hofmanninger2020dynamic,perkonigg2021dynamic}. This allows them to adapt to evolving data and new tasks. However, if models adjust too much to new information, they can lose the ability to recognize previously learned patterns, a problem known as catastrophic forgetting~\citep{kemker2018measuring}. In biomedical applications, where new disease cases are rare and accumulate slowly over time, re-training a model from scratch with each new case is inefficient and resource-intensive. Class incremental learning, which integrates new disease categories into existing models while retaining prior knowledge~\citep{vats2024incremental,sun2023few,ayromlou2024ccsi,chee2023leveraging,derakhshani2022lifelonger}, is crucial. This approach enables models to improve continuously without exhaustive re-training, efficiently handling the limited and slowly accumulating biomedical data. \autoref{fig:intro} illustrates a class-incremental learning process, where a few new tissue cell slice samples from new classes are gradually added to a pre-trained model, which was initially trained on a large set of existing classes. The model's embedding space is continuously updated as these new classes are introduced.


While current methods in class-incremental learning significantly enhance biomedical image classification performance,~\textbf{one significant problem is that they depend on numerous annotated samples for each new class.} For instance,~\citet{chee2023leveraging} introduced 625 samples per class in the CCH5000 dataset and 873 and 452 samples in two incremental steps in the EyePACS dataset. Similarly,~\citet{ayromlou2024ccsi} added thousands to tens of thousands of samples in four datasets of MedMNIST. Nevertheless, this remains unrealistic in the short term within the biomedical field. Given that acquiring and labeling biomedical images is costly and requires specialized expertise, rendering it impractical for hospitals to prepare new datasets for novel classes~\citep{feng2023learning, dai2023pfemed}. Furthermore, high data labeling costs, limited computing resources, and the rarity of certain cases pose significant challenges to deep learning and machine learning applications in the biomedical field. Consequently, \textbf{the small sample sizes of new classes lead to issues of imbalanced data distribution and long-tail data distribution between normal classes and novel classes}~\citep{singh2021metamed}. Imbalanced distribution typically refers to a dataset in which some categories are significantly more frequent than others, while long-tailed distribution~\citep{xu2022constructing} describes a severe imbalance where most categories are infrequent, with only a few categories frequently dominating the dataset. Such imbalances bias classifiers, allowing the majority of categories to substantially influence the decision boundaries~\citep{yang2020rethinking, 10114639, 10152774}.

In the current biomedical field, common strategies to address class imbalance include MixUp regularization~\citep{galdran2021balanced}, curriculum learning~\citep{jimenez2019medical}, and adjusted loss functions~\citep{galdran2020cost}, etc. However, these approaches often overlook the models' adaptation ability to the dynamic process where the number of classes increases over time, i.e., the class incremental learning capability. 

\begin{figure}[!t]
\centering
\includegraphics[width=0.4\textwidth]{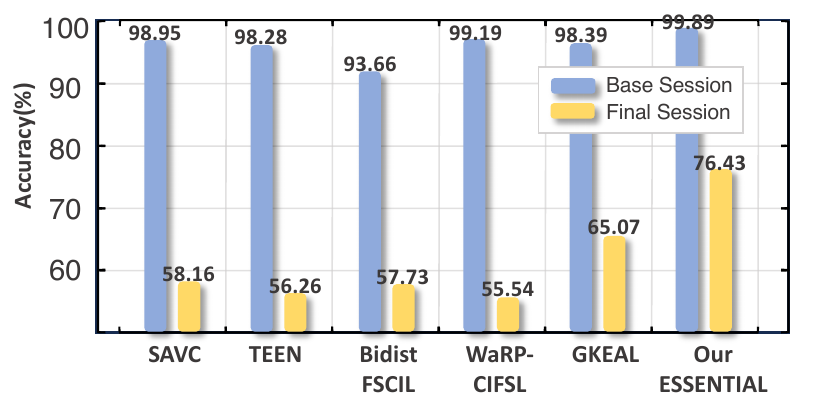}
\caption{The accuracy of the base session and the final session when applying state-of-the-art methods from CV to the biomedical dataset BloodMNIST.}
\label{fig:methods_accuracy}
\end{figure}

To the best of our knowledge, \textbf{class-incremental learning for limited samples remains underexplored in the biomedical field.} Conversely, few-shot class-incremental learning (FSCIL) has achieved remarkable success in the domain of computer vision (CV). While effective and prevalent, the unique characteristics of biomedical images, compared to natural images, pose a hurdle to directly transferring methods from the CV domain to biomedical imaging. As illustrated in \autoref{fig:methods_accuracy}, when we directly apply state-of-the-art methods analysed in \autoref{Related Work} from CV to biomedical images, the accuracy in the last session shows a significant drop compared to the base session. Therefore, one problem we need to address in the scenario is catastrophic forgetting. 

Moreover, another possible reason for the ineffectiveness of previous methods lies in that class imbalance or long-tailed data distributions can cause model bias among classes, leading to a deviation from new classes. Specifically, the introduction of new class samples, which are significantly fewer than base old class samples, results in an imbalance in magnitude within the embedding space and substantial differences in classifier weight magnitudes~\citep{hou2019learning}. Consequently, the model training tends to focus predominantly on the previously learned classes. 

As a result, we propose a novel approach ESSENTIAL (unc\textbf{E}rtainty-guided cla\textbf{SS}-incremental l\textbf{E}ar\textbf{N}ing wi\textbf{T}h l\textbf{I}mited s\textbf{A}mp\textbf{L}es via semantic expansion) in the biomedical domain for class-incremental learning under limited samples, i.e., data under class imbalance and long-tailed distribution, by addressing the problems above.
In detail, to address the catastrophic forgetting problem, we first propose a novel dynamic memory module named Uncertainty Trajectory Analyzer (UTA), for selecting and storing the most informative exemplars in each incremental session into a fixed-size memory bank. 
Therefore, the classification model can not only learn from the novel class samples but also review the representative samples stored in the memory bank. 
To avoid tracking all samples in every epoch, our prediction module effectively forecasts the cumulative entropy for all unlabeled samples and stores those with the highest cumulative entropy as exemplars. This approach tracks cumulative entropy to dynamically gauge sample uncertainty, as higher entropy reflects greater uncertainty. Furthermore, we have theoretically proved that samples with significant fluctuations in accumulated entropy, which are considered the most challenging or informative samples in UTA, have both immediate and sustained impacts on the model. 

Second, to further alleviate catastrophic forgetting, we propose a Fine-Grained Semantic Expansion Module. This module decomposes each class into more detailed sub-features, enriching the feature space and helping the model capture subtle intra-class variations. Such granular expansion of semantic details strengthens the model’s capability to consistently retain critical information across updates.

Lastly, our method mitigates classification biases rising from sample class imbalance and long-tail distributions by employing a cosine classifier, whose classification decisions are based on the direction of vectors rather than their magnitudes. We have proved that it can significantly mitigate the bias towards base classes in \autoref{fig:classifier}.

To this end, our primary contributions can be summarized as follows:
\begin{enumerate}
    \item To the best of our knowledge, we are the \textbf{first} to propose a class-incremental learning method under limited samples in the field of biomedical imaging.
    \item We propose a novel exemplar selection module named Uncertainty Trajectory Analyzer (UTA) to track and predict cumulative entropy by leveraging sample uncertainty. It can identify representative samples of each class for sample replay in each incremental session to alleviate catastrophic forgetting. We theoretically prove that cumulative entropy does indeed impact the model both immediately and over time.
    \item We propose a novel Fine-Grained Semantic Expansion module to extend the supervised contrastive learning model at a finer granularity, thereby enhancing the training context of the model.
    \item We employ a cosine classifier and empirically prove that it can significantly mitigate the classification bias towards classes of more samples. 
    \item On the same biomedical datasets PathMNIST and BloodMNIST, our method outperforms the state-of-the-art class-incremental learning approaches by a significant margin in terms of accuracy (average accuracy improvement ranged from 7.53\% to 37.12\%) on imbalanced and long-tailed datasets. Ablation studies further demonstrate the effectiveness of each proposed module.
\end{enumerate}

\section{Related Work}\label{Related Work}
\subsection{Replay-based Incremental Learning in Biomedical Area}
Incremental learning in biomedical imaging is classified into three main types: regularization-based, replay-based, and dynamic model-based methods. This discussion focuses on replay-based methods, which operate on the principle of ``learning from the past to inform the future~\citep{tian2024survey}''. During training of new tasks, these methods retain a subset of historical data to help the model recall and preserve previously learned patterns, diagnostic criteria, and treatment protocols.

The expertise from one biomedical institution can be transferred to new settings, ensuring consistent diagnostics and treatments. By preserving historical knowledge, the necessity for extensive new patient data is mitigated, reducing privacy risks during model updates~\citep{qazi2024continual}.
~\citet{zhang2022learning} combined directional and arbitrary alignment to enhance both memorization and generalization. ~\citet{bera2023memory} used SegGAN for generating data and labels for new and old tasks. ~\citet{wei2023representative} applied Variational Autoencoders with adversarial networks for data selection. ~\citet{thandiackal2024multi} proposed a multi-scale feature alignment method for unsupervised domain adaptation. Recently, ~\citet{kim2024continual} introduced a federated continual learning framework for multi-center studies without a central server.

\subsection{Few-Shot Class-Incremental Learning}
There are few studies on Few-Shot Class-Incremental Learning (FSCIL) in the biomedical area. We found only one method MAPIC~\citep{10050007}, which combines an embedding encoder, prototype enhancement module, and distance-based classifier to tackle FSCIL in biomedical time series. It efficiently handles data scarcity and prevents catastrophic forgetting, enhancing real-time health monitoring and precision decision-making systems.

Conversely, in the field of computer vision, FSCIL is a critical area with a substantial body of existing research. CEC~\citep{zhang2021few} leverages a graph-based adaptation module that incrementally adjusts classifier weights. FACT~\citep{zhou2022forward} enhances machine learning models by pre-allocating embedding space with virtual prototypes during initial training, allowing for seamless integration of new class data. CLOM~\citep{zou2022margin} addresses the dilemma of balancing base and novel class performance by applying additional constraints to margin-based learning, significantly enhancing both the discriminability and generalization capabilities. SAVC~\citep{song2023learning} also utilizes virtual classes to enhance class separation and enable robust learning of new and existing classes. TEEN~\citep{wang2023fewshot} improves the recognition of new classes by calibrating new prototypes with weighted base prototypes, enhancing classification without additional training. BidistFSCIL~\citep{zhao2023few} uses dual knowledge sources—rich general knowledge from a base model and adapted knowledge from an updated model—to simultaneously prevent forgetting of old classes and overfitting to new classes. WaRP-CIFSL~\citep{kim2023warping} proposes two main strategies: first, using a robustly trained base model to maintain the stability of past knowledge; second, adopting a dynamically updated model to adapt and learn new class information. GKEAL~\citep{Zhuang_GKEAL_CVPR2023} employs a Gaussian Kernel Embedded Analytic module that transforms network training into a linear problem, supporting phase-based recursive learning. This approach enables seamless integration of new class data without forgetting existing categories.

\subsection{Exemplar Selection Method}
Exemplar selection strategies are crucial for enhancing machine learning model efficiency by identifying the most beneficial training samples. Traditional methods include Selection Via Proxy, which employs smaller, proxy models to expedite data selection by predicting the utility of larger, more complex datasets~\citep{coleman2019selection}. Another approach, utilized in~\citep{rebuffi2017icarl}, involves selecting exemplars based on the nearest-mean-of-exemplars (NME) method, where each class prototype is represented by the mean of its exemplar feature vectors. Meanwhile,~\citet{chhabra2024data} adopted a strategy that leverages influence functions to identify and eliminate training samples that negatively impact model performance.

Additionally, active learning strategies are designed to identify and annotate the most informative samples from an unlabeled data pool. Techniques vary from pool-based~\citep{nath2020diminishing,zhang2020state,zhan2021comparative}, which selects samples from a large pool are labeled, to stream-based~\citep{saran2023streaming,cacciarelli2024active,martins2023meta}, which makes real-time decisions on incoming data, and committee-based~\citep{zhao2024committee}, which relies on model disagreement to assess data informativeness.


\section{Background Theory and Motivation}
\subsection{Characteristics of Biomedical Images}
Biomedical imaging has distinct characteristics that differentiate it from natural images~\citep{wang2023review}. These images focus on small, specific regions of interest (ROIs), such as microaneurysms or consolidations in X-rays, which are often difficult to detect due to subtle texture variations~\citep{wilkinson2003proposed,Debiasing_VQA}. Unlike the rich color and content diversity found in natural images, biomedical images typically exhibit a monochromatic color scheme with less diversity and contrast. Additionally, biomedical image datasets are relatively small, ranging from one thousand~\citep{khosravi2018robust} to one hundred thousand images~\citep{gulshan2016development, rajpurkar2017chexnet}, compared to the millions in datasets like ImageNet. This scarcity, combined with high variability in visual attributes, presents unique challenges in the analysis of biomedical images. Accurately quantifying uncertainty in these images is crucial for identifying ambiguities, artifacts, and novel patterns~\citep{ghesu2021quantifying, linmans2023predictive}, while efficient feature representation is essential for reducing dimensionality and extracting critical biomedical information.
\textbf{Thus, exemplars should be diverse and well-distributed throughout the data \citep{wang2024comprehensive}.}

\subsection{Motivations and Insights}
\noindent
\textbf{Severe Catastrophic Forgetting for Previous Methods}\ : 
As mentioned in the \autoref{introduction} in the field of biomedical image classification, there is currently no class-incremental method specifically designed to address imbalanced and long-tailed data distributions in the context of continuously evolving, limited sample data streams. When methods from FSCIL in CV are directly applied to biomedical images, as shown in \autoref{fig:methods_accuracy} and \autoref{tab:sota}, the classification accuracies decrease drastically, indicating that severe catastrophic forgetting occurs.

\noindent
\textbf{Representation Overlapping Caused by Supervised Contrastive Learning}\ : 
Supervised Contrastive Learning (SCL) enhances intra-class consistency by clustering same-class samples tightly in embedding space and mitigates inter-class overlap by pushing different-class samples apart, boosting discriminative power~\citep{mai2021supervised,peng2023sclifdsupervisedcontrastiveknowledgedistillation}. Combining SCL with CE loss better balances class separation and clustering.
While SCL slightly improves class separation, it implicitly reduces inter-class distance, causing inevitable overlap between novel and old classes in the representation space~\citep{song2023learning}. Our experiments also confirm this issue, as demonstrated in \autoref{fig:network} (a). Given the inherent properties of biomedical images—complex, noisy~\citep{budd2021survey}, and featuring small, monochromatic regions that lack the diversity and contrast of natural images—this often results in more severe overlap.
To address this, our Fine-Grained Semantic Expansion module generates numerous augmented virtual samples for each class in both base and incremental stages, which can be considered finer-grained subclasses derived from the original classes. Their semantic information fills the current class's representation space, leaving sufficient room for future updates and mitigating inevitable overlaps between novel and old classes. As presented in \autoref{fig:network}, model with semantic expansion exhibits more compact intra-class distributions and more dispersed inter-class separations.

\noindent
\textbf{Selective Sample Replay Strategy}\ :
Previously we have introduced that sample reply is conducive to alleviating catastrophic forgetting as the model can review the representative sample of the old classes in each incremental session.
But replaying all samples from previous classes is impractical due to the significant computational and storage resource requirements. This raises the question: how can we select the most representative samples for each class and replay their information in a compact form to subsequent models?

Traditional data selection strategies, such as NME~\citep{rebuffi2017icarl}, and certain classic active learning methods typically focus on static model information at the end of each training session. As shown in \autoref{fig:uncertainty_epoch}, samples selected using these strategies often lead to a slower convergence of model uncertainty, resulting in lower learning efficiency and poorer predictive performance in the final model state.
Therefore, we use the model's dynamic behavior on individual samples as a measure of sample uncertainty. Prior research has predicted the training dynamics of unlabeled data to estimate uncertainty~\citep{kye2023tidal} and traced the dynamic loss of each sample during training~\citep{wan2024tracing}. These studies interpret sample uncertainty measurement from the perspectives of fine-grained checkpoints and coarse-grained periods, respectively.
Building upon these studies, our UTA continuously captures the cumulative change in entropy as model parameters are updated through gradient descent. As the model iteratively minimizes the loss function and adjusts its parameters, UTA quantifies these updates by integrating the cumulative changes in entropy
and selects the samples with the highest cumulative entropy as the most informative ones to update the memory bank. 

\noindent
\textbf{Classification Bias Resulting from Class Imbalance}\ :
Existing class-incremental learning methods in the biomedical field often rely on a large number of labeled samples, which is not consistent with real-world scenarios (illustrated in \autoref{fig:intro}). The number of samples in the base classes is significantly larger than that of the newly introduced, imbalanced or long-tailed incremental classes. In traditional classifiers, the feature vector lengths of base classes in the embedding space are much greater than those of the incremental classes, leading to a bias in the classifier's weights toward the base classes. Our cosine classifier normalizes feature vectors to a unit sphere, emphasizing their direction rather than magnitude. As depicted in \autoref{fig:classifier}, cosine classifier effectively prevents weight bias toward the base classes compared to other classifiers, significantly reducing the misclassification of new samples into base classes.

\subsection{Supervised Contrastive Learning}
Contrastive learning enhances classifier performance by promoting intraclass compactness—alignment within classes—and interclass separability—distinction between classes~\citep{khosla2020supervised}.

The basic principle of self-supervised contrastive learning involves bringing each anchor closer to a positive sample while distancing it from negative samples within the embedding space~\citep{khosla2020supervised}. In our method, we incorporate the MoCo~\citep{chen2020improved, he2020momentum} framework to enhance traditional supervised contrastive learning. For each labeled sample $(x,y)$ from a batch size $N$, the query augmentation $x^{(q)}=Aug_{q}(x)$ and key augmentation $x^{(k)}=Aug_{k}(x)$ are transformed by query network $g_{q}$ and key network $g_{k}$ into L2-normalized vectors $r^{(q)}=\frac{h(f(x^{(q)}))}{\|h(f(x^{(q)}))\|}$ and $r^{(k)}=\frac{h(f(x^{(k)}))}{\|h(f(x^{(k)}))\|}$, respectively. Both $g_{q}$ and $g_{k}$ utilize the same architectures for the feature extractor $f$ and the projector $h$, ensuring consistency in how features are processed. The key difference is that $g_{k}$ updates its parameters more slowly than the query network, employing a momentum mechanism defined by $\theta_k^{(t+1)}=\mu\theta_k^{(t)}+(1-\mu)\theta_q^{(t)}$, where $\mu$ is the momentum coefficient. This setup supports stability in learned features over time. These transformations yield a ``multiviewed batch'' comprising $2N$ augmented instances, denoted as $\{(\tilde{x}_\ell,\tilde{y}_\ell)\}_{\ell=1}^{2N}$, where $\text{ }\tilde{y}_{2k-1}=\tilde{y}_{2k}=y_{k}$ ensures label consistency across the views for each original sample. Furthermore, we uphold synchronized feature and label queues, denoted as $\mathcal{Q}_{feature}$ and $\mathcal{Q}_{label}$ respectively, each of identical length, to retain the latest key embeddings $k$ and their associated labels. The supervised contrastive loss for a given query vector $r^{(q)}$ in relation to the batch of augmented instances is then calculated by:
\begin{equation}
L_{\text{SCL}} = -\frac{1}{|\mathcal{P}(r^{(q)})|} \sum_{r^{(k^+)} \in \mathcal{P}(r^{(q)})} \log \left( \frac{\exp((r^{(q)} \cdot r^{(k^+)}) / \tau)}
{\sum_{r^{(k')} \in \mathcal{A}(r^{(q)})} \exp((r^{(q)} \cdot r^{(k')}) / \tau)} \right)
\label{eq:SCL}
\end{equation}
In this formula, $\mathcal{P}(r^{(q)})$ denotes the set of positive examples sharing the same class as $r^{(q)}$, and $\mathcal{A}(r^{(q)})$ represents the set of all augmented examples in the current batch. The temperature parameter $\tau$ is used to scale the dot product of embeddings, aiding in the stabilization of gradients during training. This loss structure promotes the proximity of similar class embeddings while enhancing the separation between different classes.

\subsection{Cosine Similarity}
Class imbalance or long-tailed distributions can cause bias among classes, leading to deviations from new classes. Specifically, the introduction of new class samples, which are far fewer than base class samples, results in an imbalance in the embedding space and significant differences in classifier weights, causing the model to focus on previously learned classes. Cosine similarity, defined as:
\begin{equation}
\label{sim}
    \mathrm{sim}(x,y)=\frac{x^Ty}{\|x\|\|y\|},
\end{equation}
focuses on the angular relationship between vectors, effectively mitigating this bias by normalizing the embeddings ~\citep{hou2019learning}. Now, the probability of a sample $x$ belonging to prototype $j$ is calculated as:
\begin{equation}
    y_i^\star=\frac{\exp(\eta\langle p_j,f_\theta\left(x_i\right))}{\sum_j\exp(\eta\langle p_k ,f_\theta\left(x_i\right))},
\end{equation}
where $p_j$ and $f_\theta\left(x_i\right)$ are the $l_2$-normalized prototype embedding and feature vector, respectively, and $\eta$ controls the sharpness of the distribution.

By emphasizing direction over magnitude, cosine similarity normalizes vectors to a unit sphere to ensure fair and consistent comparison between new and existing classes.

\begin{figure*}[!t]
\begin{center}
\includegraphics[width=1.0\textwidth]{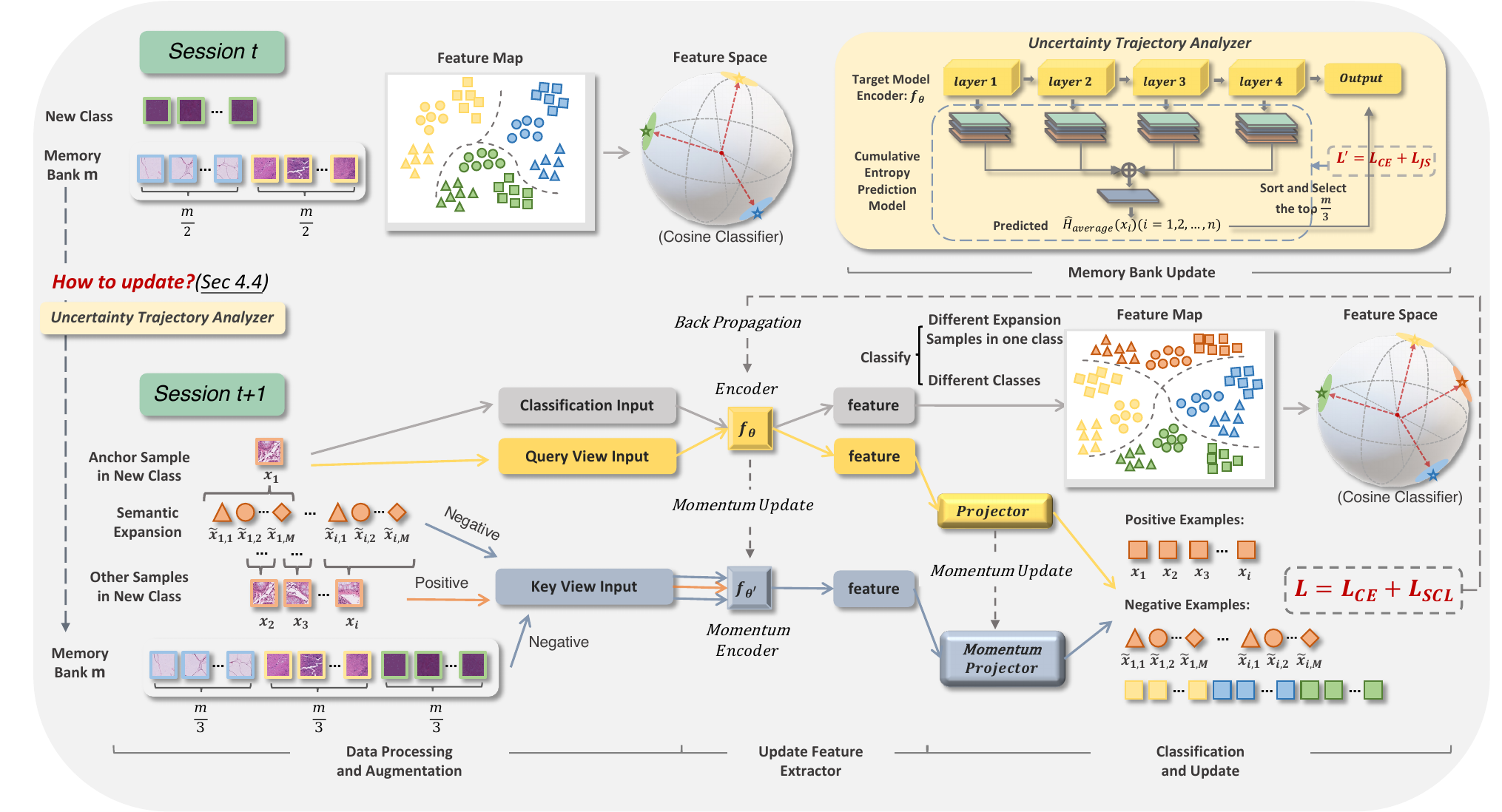}
 \caption{\textbf{Overview} of our proposed \textbf{ESSENTIAL}. Target model includes a classification and supervised contrastive task, with shared backbone for the classification and contrastive query networks. The contrastive key network is updated via momentum. A cosine classifier distinguishes between classes and semantic expansions. Supervised contrastive loss and classification loss jointly update the model. Besides, The memory bank is updated by an additional module ``Uncertainty Trajectory Analyzer'', selecting the most uncertain samples as exemplars.  }
 \label{fig:pipeline}
 \end{center}
 \vspace{-15pt}
\end{figure*}

\section{Methodology}

\subsection{Problem Formulation}
In the scenario of our class-incremental learning under limited tail instances, we define ``limited data'' to include both ``imbalanced data'', represented by 50 samples per incremental class, and ``long-tailed data'', epitomized by 20 samples per class, grounded in the need to simulate real-world data distributions, typically characterized by long tails~\citep{openlongtailrecognition}. 

The objective of class-incremental learning under limited tail instances is to acquire new classes from limited training samples while preserving the performance on previously learned classes. In practical settings, it is often implemented in a sequence of learning sessions, the cumulative training dataset including a limited number of examples from newly introduced classes denoted as $\mathcal{D}_{\text{train}} = \{\mathcal{D}_{\text{train}}^{(0)}, \mathcal{D}_{\text{train}}^{(1)}, \ldots, \mathcal{D}_{\text{train}}^{(T)}\}
$ across sessions $t=0$ to $T$, where $T$ is the total number of sessions. Each session $t$ contains a training dataset $\mathcal{D}_{\text{train}}^{(t)} = \{(x_i, y_i) \mid i = 1, \ldots, N_t\}$, with $x_i\in\mathbb{R}^D$ representing the 
$i$-th image and $y_i\in\mathcal{C}^{(t)}$ the corresponding label from label space $\mathcal{C}^{(t)}$. The test dataset for session 
$t$, denoted as $\mathcal{D}_{\mathrm{test}}^{(t)}$, evaluates the model on the union of all classes up to session $\mathcal{D}_{\mathrm{test}}^{(t)}\supseteq\bigcup_{j=0}^t\mathcal{C}^{(j)}$, where each $\mathcal{C}^{(j)}$ represents the label space for session $j$. Each session introduces non-overlapping class labels, $i.e., \mathcal{C}^{(i)}\cap\mathcal{C}^{(j)}=\emptyset$, $\forall i\neq j$. The model begins with an initial dataset $D^{(0)}$ that includes a diverse set of classes $C^{(0)}$, providing ample examples for each class. For each following incremental session $t$, $D^{(t)}$ for $t>0$ contains limited labeled samples.

\subsection{Method Overview}

The complete algorithm workflow and pipeline are presented in \autoref{fig:pipeline}. The target model includes a classification task and a supervised contrastive task, with the classification network and contrastive query network sharing the same backbone. The contrastive key network is updated via momentum, with its parameters partially aligned to the query network during each iteration. The classification task uses a Cosine Classifier(\autoref{Cosine Classifier}) to distinguish both between different classes and between different semantic expansions within the same class. The anchor sample generates new samples through Fine-grained Semantic Expansion(\autoref{Denser Semantic Space}), with their embeddings and those of previous classes in the memory bank forming its negative sample set, while embeddings from the same class serve as its positive sample set. Overall, the supervised contrastive loss $L_{\text{SCL}}$ and classification loss $L_{\text{CE}}$ jointly update the model. Ultimately, the memory bank is updated using the Uncertainty Trajectory Analyzer(\autoref{Exemplar Selection}), selecting the most uncertain samples (with highest cumulative entropy) as exemplars for storage.

\subsection{Fine-Grained Semantic Expansion}
\label{Denser Semantic Space}
We generates virtual samples through transformations such as rotation and color adjustment of original biomedical images, artificially expanding the dataset to increase sample diversity. This strategy enables the model to learn richer semantic information and more robust feature representations. These virtual samples not only serve as a rehearsal for new classes, enhancing the model's adaptability to unseen classes, but also act as carriers of semantic knowledge, providing a more comprehensive learning context for the model.

Inspired by the idea from ~\citet{song2023learning} and ~\citet{lee2020self} of creating virtual samples of the base class to fill the unallocated embedding space, our approach performs semantic expansion on all samples, whether they appear in base session or incremental session, i.e., the original samples are subjected to a series of transformations (e.g., rotations, flips, etc.) to generate a sequence of virtual samples to supplement the original semantic space. This process begins with the application of a predefined set of transformations $T=\{t_1,t_2,...,t_M\}$, to each sample $x$ in exemplar set. These transformations yield a collection of augmented instances $\{\tilde{x}_{i,1},\tilde{x}_{i,2},...,\tilde{x}_{i,M}\}$, where each $\tilde{x}_{i,j}=t_j(x_i)$ corresponds to the $j$-th transformation applied to the $i$-th sample. All transformed instances $\tilde{x}_{i,j}$ are then fed into the neural network along with the original samples to obtain their feature embeddings $\tilde{z}_{i,j}$, which are evaluated by $f_\theta(\tilde{x}_{i,j})$.

As incremental learning progresses, both the feature extractor and classifier are dynamically updated in a fine-grained manner. Next, we further refine the new class prototypes in each session in session $t$ as :
\begin{equation}
\label{eq}
P^{\prime}=\left\{
\begin{array}{ll}
\begin{aligned}
&\{\tilde{p}_{1,1}^{t},\tilde{p}_{1,2}^{t}, \ldots, \tilde{p}_{1,M}^{t}\} \cup \ldots \cup \{\tilde{p}_{m,1}^{t}, \tilde{p}_{m,2}^{t}, \ldots, \tilde{p}_{m,M}^{t}\} \\
& \hspace{5cm} \text{for } t = 0 \ldots t-1, \\
&\{\tilde{p}_{1,1}^{t},\tilde{p}_{1,2}^{t}, \ldots, \tilde{p}_{1,M}^{t}\} \cup \ldots \cup \{\tilde{p}_{|c^{(t)}|,1}^{t}, \tilde{p}_{|c^{(t)}|,2}^{t}, \ldots, \tilde{p}_{|c^{(t)}|,M}^{t}\} \\
& \hspace{5cm} \text{for } t = t
\end{aligned}
\end{array}
\right.
\end{equation}

From a finer granularity perspective, prototypes that share the same transformations can be grouped into a new prototype set. By considering multiple angles or transformed versions during prediction, model selects the class represented by the prototype whose overall feature representation is most similar to all transformed versions of the sample $x_i$:
\begin{equation}
    y_i^\star=\operatorname*{argmax}_{c\in\cup_{t=0}^{t}\mathcal{C}^{(t)}}\sum_{m=1}^{M}sim\left(f_\theta\left(\tilde{x}_{i,m}\right),\tilde{p}_{c,m}^{t}\right)
\end{equation}

Within this setup, two classifiers operate on the extracted features: Class Classifier($\psi$), parameterized by $\Phi $ is tasked with accurately predicting the class label of each transformed sample, effectively discerning between different object categories ($\hat{y}_{i,j}=\psi(\tilde{z}_{i,j};\Phi$). Simultaneously, Transformation Classifier($\phi$), utilizing parameters $\Gamma $, identifies the type of transformation applied to the sample ($\hat{t}_{i,j}=\phi(\tilde{z}_{i,j};\Gamma)$). Hence, in a fully supervised setting, we accomplish optimization based on a multi-task learning framework utilizing two different losses for optimizing the primary task and the self-supervised task in a shared feature space:
\begin{equation}
\begin{split}
L_{MT}(x_i, y_i ; \Theta, \Phi, \Gamma) &= \frac{1}{M} \sum_{j=1}^M \Big[ L_{CE}(\psi(\tilde{z}_{i, j} ; \Phi), y_i) \\
&\quad + L_{CE}(\phi(\tilde{z}_{i, j} ; \Gamma), j) \Big]
\end{split},
\end{equation}
where $L_{CE}$ is the cross-entropy loss defined in~\autoref{eq:CE}. This cohesive operation ensures the model not only learns to classify the object (what) but also understands the nature of its transformation (how).

The embedding space becomes denser and contains richer semantic information. Consequently, our supervised contrastive loss~\autoref{eq:SCL} can be refined as follows:
\begin{equation}
\begin{aligned}
L_{SCL}(g; x, \Theta, \Phi) = & -\frac{1}{N} \sum_{i=1}^N \frac{1}{M} \sum_{j=1}^M \\
& \log \left( \frac{\exp \left(\langle \phi(\tilde{x}_{i,j}; \Theta), \phi(\tilde{x}_{i,j}; \Theta) \rangle / \tau\right)}{\sum_{k=1}^M \sum_{l \neq j} \exp \left(\langle \phi(\tilde{x}_{i,j}; \Theta), \phi(\tilde{x}_{i,l}; \Theta) \rangle / \tau\right)} \right)
\end{aligned}
\end{equation}

\begin{figure}[!t]
\centering
\includegraphics[width=0.5\textwidth]{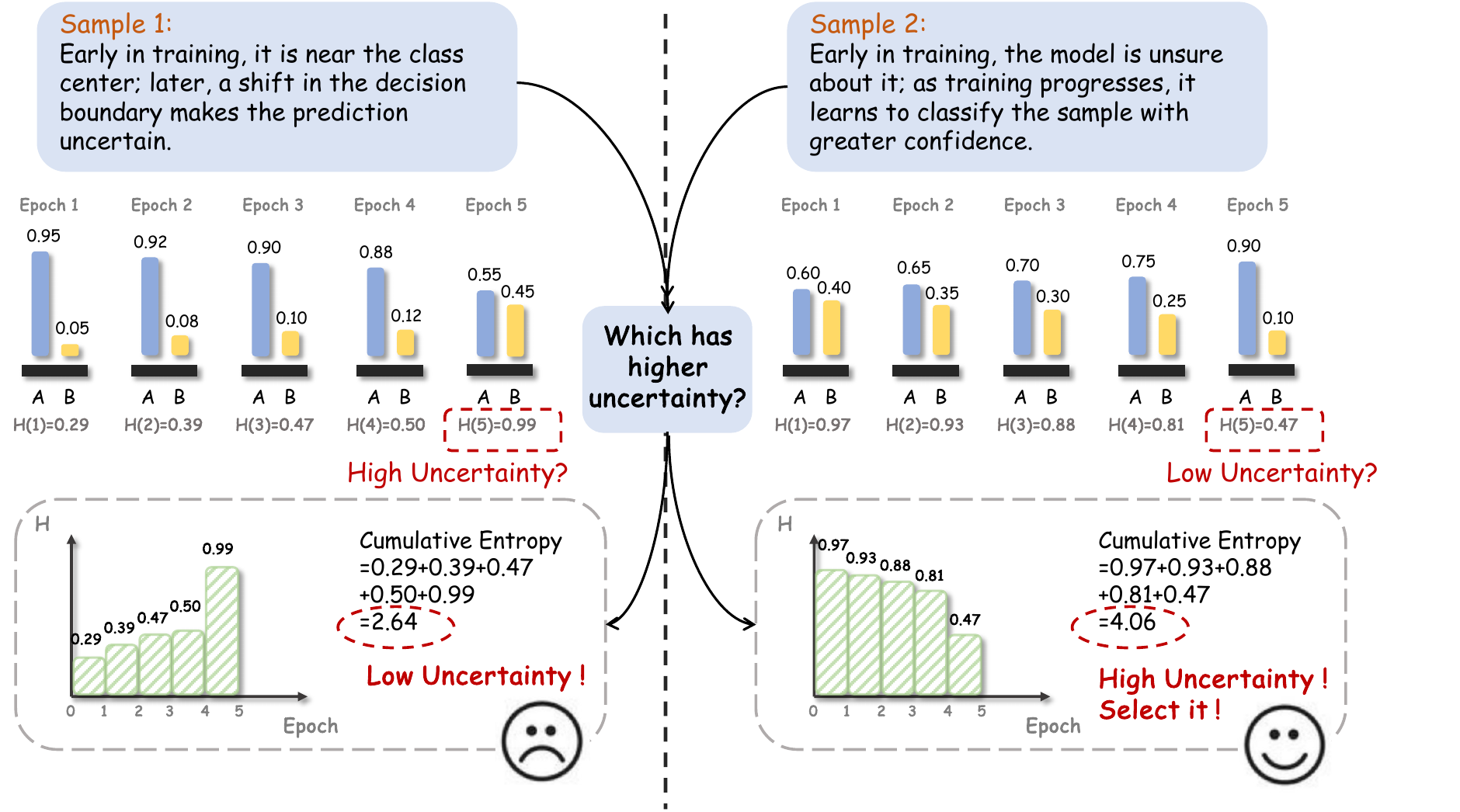}
\caption{The figure depicts a binary classification task with only 5 epochs. Sample 1 and Sample 2 represent two typical cases from a real training process, showing the probability distribution for each class at every epoch. Unlike the static approach of calculating entropy using only the final epoch’s probability distribution, we use cumulative entropy over the entire training period to assess the uncertainty of the samples.}
\label{fig:uncertainty}
\end{figure}

\begin{figure}[!t]
\centering
\includegraphics[width=0.5\textwidth]{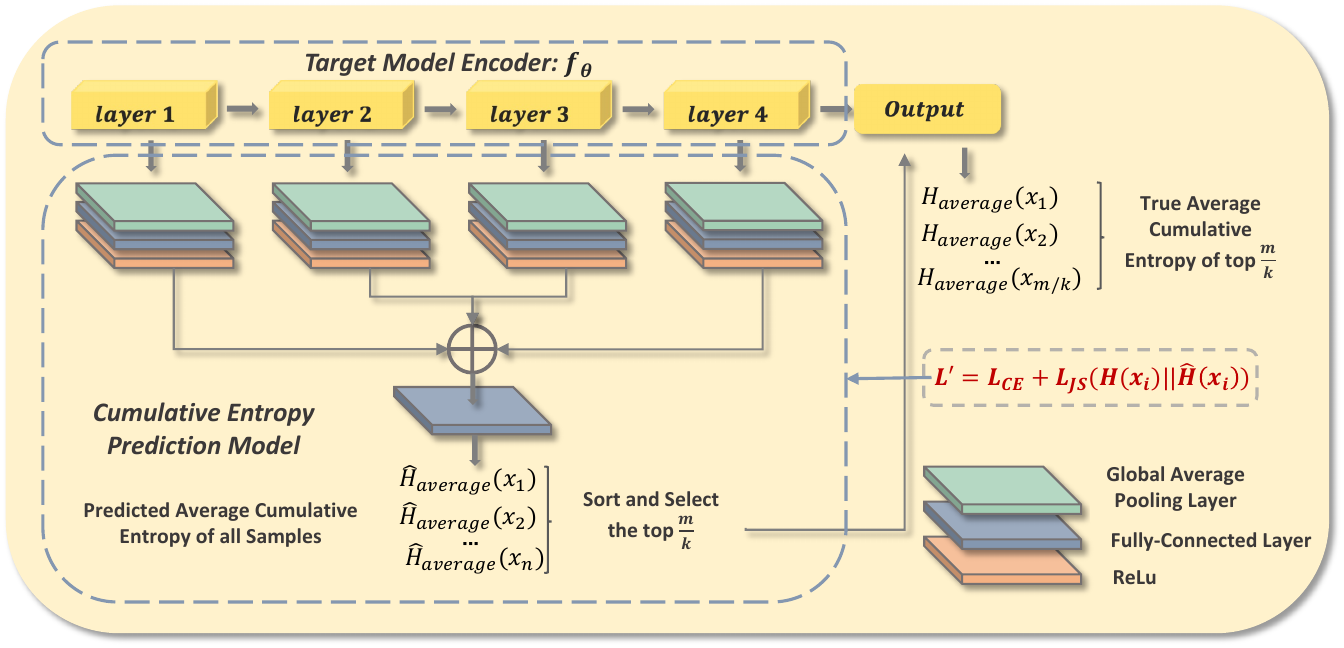}
\caption{Architecture of Cumulative Entropy Prediction Module. Multi-layer features are extracted from each intermediate layer of the target model and used as inputs to the prediction model. These features are dimensionally reduced and concatenated before being passed through the next fully-connected layer, which outputs the predicted average cumulative entropy. The predicted average cumulative entropy of all samples within a class is then sorted, and the top samples are selected for re-evaluation by the target model to compute their true average cumulative entropy. Finally, the prediction model is updated using the Jensen-Shannon divergence between the predicted and true entropies, along with the cross-entropy loss of the target model.}
\label{fig:UTA}
\end{figure}


\subsection{Uncertainty Trajectory Analyzer}
\label{Exemplar Selection}

\subsubsection{Defining Cumulative Entropy}
In information theory, entropy $H$ is utilized to quantify the uncertainty present within a probability distribution. The static entropy of each sample $x$ at time $t$ is calculated based on the probability distribution $p(c|x)$ ($c$ is the class label) assigned by the model:
\begin{equation}
\label{entropy}
    H^{(t)}(x)=-\sum_{c=1}^Cp_t(c|x)\log p_t(c|x)
\end{equation}

We define cumulative entropy as the approximate area under the entropy curve throughout the entire training period, recognizing that the training epochs are discrete. This is mathematically represented as the sum of areas of trapezoids formed between successive epochs. It is expressed as:
\begin{equation}
    H_{cumulative}=\int_0^TH(t)dt\approx\sum_{t=1}^{T-1}\frac{H(t)+H(t+1)}2\Delta t, 
\end{equation}
where $H(t)$ represents the entropy of the model at time $t$, $\Delta t$ is the duration of each epoch, and $T$ is the total duration of the training period.

To obtain the cumulative entropy for each sample, we start by determining the predicted probability distribution for each sample $x_i$ at each time step $t$, denoted as $p^{(t)}(c|x_i)$.
\begin{equation}
    \mathbf{p}^{(t)}(x_i)=\left[p^{(t)}(1|x_i),p^{(t)}(2|x_i),\ldots,p^{(t)}(C|x_i)\right]
\end{equation}
After plugging into ~\autoref{entropy}, we get the entropy of $x_i$:
\begin{equation}
    H^{(t)}(x_i)=-\sum_{c=1}^Cp^{(t)}(c|x_i)\log p^{(t)}(c|x_i)
\end{equation}

Next, we calculate the cumulative entropy of $x_i$ over the entire training period, which is the sum of the entropy values over all time points:
\begin{equation}
    H_{cumulative}(x_i)=\sum_{t=1}^TH^{(t)}(x_i)=-\sum_{t=1}^T\sum_{c=1}^Cp^{(t)}(c|x_i)\log p^{(t)}(c|x_i),
\end{equation}
and the average cumulative entropy of $x_i$ is:
\begin{equation}
    H_{average}(x_i)=\frac1TH_{cumulative}(x_i)=\frac1T\sum_{t=1}^TH^{(t)}(x_i)
\end{equation}

We illustrate the difference between using the probability distribution from the final epoch to calculate a sample’s entropy and the sample’s cumulative entropy over the entire training period through the example in \autoref{fig:uncertainty_epoch}.
\begin{table*}[htbp]
\scriptsize
\centering
\caption{Statistical Information of Dataset.}
\label{tab:dataset}
\begin{tabular}{@{}cccccccccc@{}}
\toprule
Dataset                     & \begin{tabular}[c]{@{}c@{}}Data\\ Composition\end{tabular} & \begin{tabular}[c]{@{}c@{}}Total \\ Classes\end{tabular} & \begin{tabular}[c]{@{}c@{}}Number of \\ Sessions\end{tabular} & \begin{tabular}[c]{@{}c@{}}Number of \\ Classes in \\ Base Session\end{tabular} & \begin{tabular}[c]{@{}c@{}}Number of \\ Class in Each\\ Incremental Session\end{tabular} & \begin{tabular}[c]{@{}c@{}}Number of \\ Samples per \\ Base Class\end{tabular} & \begin{tabular}[c]{@{}c@{}}Number of \\ Samples per\\ Incremental Class\end{tabular} & \begin{tabular}[c]{@{}c@{}}Memory\\  Size\end{tabular} & Resolution               \\
\midrule
\multirow{2}{*}{PathMNIST}  & Imbalanced                                                 & \multirow{2}{*}{9}                                       & \multirow{2}{*}{7}                                               & \multirow{2}{*}{3}                                                              & \multirow{2}{*}{1}                                                                       & \multirow{2}{*}{1000}                                                          & 50                                                                                   & 200                                                    & \multirow{2}{*}{28*28}   \\
                            & Long-tailed                                                &                                                          &                                                                  &                                                                                 &                                                                                          &                                                                                & 20                                                                                   & 70                                                     &                          \\
\multirow{2}{*}{BloodMNIST} & Imbalanced                                                 & \multirow{2}{*}{8}                                       & \multirow{2}{*}{7}                                               & \multirow{2}{*}{2}                                                              & \multirow{2}{*}{1}                                                                       & \multirow{2}{*}{800}                                                           & 50                                                                                   & 150                                                    & \multirow{2}{*}{224*224} \\
                            & Long-tailed                                                &                                                          &                                                                  &                                                                                 &                                                                                          &                                                                                & 20                                                                                   & 60                                                     &                          \\
\bottomrule
\end{tabular}%
    \begin{tablenotes}
        \centering
        \footnotesize    
        \item[1]  In imbalanced scenarios, sample sizes should range from 21 to 50, and in long-tailed scenarios, from 0 to 20.~\citep{openlongtailrecognition}
    \end{tablenotes}
\end{table*}

\subsubsection{Theoretical Proof}
To demonstrate that samples with the maximum cumulative entropy, defined as the area under the curve of sample entropy at the final timestep, are more uncertain or have a greater impact on the model, we provide proofs from two perspectives: the immediate impact and the long-term influence.

\textbf{Immediate Impact.} The gradient of entropy defined in \autoref{entropy} with respect to the model parameters $\theta $ is:
\begin{equation}
    \nabla_\theta H^{(t)}(x)=-\sum_{c=1}^C\left(\frac{\partial p_t(c|x)}{\partial\theta}\log p_t(c|x)+p_t(c|x)\frac{\partial\log p_t(c|x)}{\partial\theta}\right)
\end{equation}
This gradient captures how the entropy changes as the model parameters are updated during training.

Then, we introduce influence function providing an approximation of how a single sample $x$ affects the model parameters $\theta $. Specifically, the change in the model parameters $\Delta\theta $ due to the inclusion of sample $x$ is given by:
\begin{equation}
    \Delta\theta=-H_\theta^{-1}\nabla_\theta L(\theta_t,x),
\end{equation}
where $H_\theta^{-1}$ is the inverse of the Hessian matrix of the loss function with respect to the model parameters $\theta $, and $\nabla_\theta L(\theta_t,x)$ is the gradient of the loss function at time $t$ for sample $x$.

Ultimately, we combine the entropy gradient $\nabla_\theta H^{(t)}(x)$ with the influence function to determine the change in entropy due to the influence of $x$:
\begin{equation}
    \Delta H^{(t)}(x)\approx\nabla_\theta H^{(t)}(x)\cdot\Delta\theta\approx\nabla_\theta H^{(t)}(x)\cdot\left(-H_\theta^{-1}\nabla_\theta L(\theta_t,x)\right)
\end{equation}
To capture the cumulative effect of this influence over the entire training period, it is integrated over time from 0 to $T$, leading to the cumulative entropy $\mathrm{CE}_I(x)$:
\begin{equation}
    \mathrm{CE}_I(x)=\int_0^T\left(\nabla_\theta H^{(t)}(x)\cdot\left(-H_\theta^{-1}\nabla_\theta L(\theta_t,x)\right)\right)dt
\end{equation}
On the one hand, the gradient of entropy $\nabla_\theta H^{(t)}(x)$ reflects the sensitivity of a sample's entropy to the model parameters $\theta $: when small changes in the model parameters significantly impact the entropy of a sample $x$ it indicates that the model's predictions for this sample are highly unstable, and the model lacks sufficient confidence in this sample. One the other hand, the influence function $H_\theta^{-1}\nabla_\theta L(\theta_t,x)$ essentially measures a sample's contribution to the overall adjustment of the model parameters, which indirectly affects the model's prediction uncertainty for other samples. Therefore, samples with larger influence functions are more likely to contribute significantly to the model's overall uncertainty.

\textbf{Long-term Influence.} To understand how entropy changes as a function of time, influenced by the changing model parameters $\theta $, we take the derivative of entropy using chain rule:
\begin{equation}
\label{derivative of entropy}
    \frac{dH}{dt}=-\sum_{c=1}^C\left(\frac{\partial p(c|x)}{\partial\theta}\cdot\frac{d\theta}{dt}\right)\log p(c|x)+\frac{p(c|x)}{p(c|x)}\cdot\left(\frac{\partial p(c|x)}{\partial\theta}\cdot\frac{d\theta}{dt}\right)
\end{equation}

We assume that the model parameters $\theta $ are updated using gradient descent. The update rule for the parameters at time $t$ is given by:
\begin{equation}
    \frac{d\theta}{dt}=-\eta\nabla_\theta L(\theta),
\end{equation}
here $\eta$ is the learning rate and $\nabla_\theta L(\theta)$ is the gradient of the loss function.
Substituting this into \autoref{derivative of entropy}:
\begin{equation}
    \frac{dH}{dt}=\sum_{c=1}^C\frac{\partial p(c|x)}{\partial\theta}\cdot\eta\nabla_\theta L(\theta)\cdot(1+\log p(c|x))
\end{equation}
Integrating the entropy change over the entire training period from 0 to $T$, we derive the cumulative entropy $\mathrm{CE}(x)$:
\begin{equation}
    \mathrm{CE}(x)=\int_0^T\left(\int_0^t\sum_{c=1}^C\frac{\partial p(c|x)}{\partial\theta}\cdot\eta\nabla_\theta L(\theta)\right.\cdot\left(1+\log p(c|x)\right)d\tau\Bigg)dt
\end{equation}
The inner sum over class $c$ captures the sensitivity of parameter changes on the predicted probabilities of sample $x$ across all classes, integrating from 0 to $t$ to reflect the cumulative impact of these updates on predictive uncertainty. The outer integral from 0 to $T$ accumulates these effects over the entire training process.

\subsubsection{Cumulative Entropy Prediction Module}
Calculating the cumulative entropy for each sample individually is highly time-consuming and resource-intensive. Inspired by~\citep{yoo2019learning}, we employ an additional module to approximate the definition of cumulative entropy for predicting the cumulative entropy of newly introduced samples. Initially, to utilize more granular information, we extract multiple multi-scale feature maps from several hidden layers of the target model as inputs to the module. These feature maps undergo global average pooling and fully connected layers, where they are dimensionally standardized and merged. After the integration of multi-layer features, they are mapped through a fully connected layer into predictions of a multi-dimensional sample probability distribution, upon which the cumulative entropy is calculated. This module operates in parallel to the target model, requires fewer parameters, and its architecture is delineated in \autoref{fig:UTA}.

Over $T$ epochs, the model generates a predicted probability matrix $Q$ for each sample $x_i$:
\begin{equation}
    \hat{Q}^{(i)}=\begin{bmatrix}
    \hat{p}^{(1)}(1|x_{i}) & \hat{p}^{(1)}(2|x_{i}) & \cdots & \hat{p}^{(1)}(C|x_{i})\\
    \hat{p}^{(2)}(1|x_{i}) & \hat{p}^{(2)}(2|x_{i}) & \cdots & \hat{p}^{(2)}(C|x_{i})\\
    \vdots & \vdots & \ddots & \vdots\\
    \hat{p}^{(T)}(1|x_{i}) & \hat{p}^{(T)}(2|x_{i}) & \cdots & \hat{p}^{(T)}(C|x_{i})
    \end{bmatrix}
\end{equation}
Here, each row $\hat{p}^{(j)}$ represents the predicted probabilities for sample $x_{i}$ at the $j$-th epoch. Subsequently, we can obtain the predicted cumulative entropy $\hat{H}_{cumulative}(x_i)$ and the average cumulative entropy $\hat{H}_{average}(x_i)$, respectively.


By minimizing the Jensen-Shannon divergence between the actual and predicted average cumulative entropy, we establish the loss function for the prediction module. To apply it, the average distribution $M$ between the actual $H_{average}(x_i)$ and the predicted CE $\hat{H}_{average}(x_i)$ is defined as:
\begin{equation}
    M=\frac12(H_{average}(x_i)+\hat{H}_{average}(x_i))
\end{equation}
The JS divergence is then calculated as:
\begin{equation}
\begin{aligned}
    \mathcal{L}_{\mathrm{JS}}(H_{average}(x_i) \parallel \hat{H}_{average}(x_i)) = \frac{1}{2} \biggl[ &\mathcal{L}_{\mathrm{KL}}(H_{average}(x_i) \parallel M) \\
    &+ \mathcal{L}_{\mathrm{KL}}(\hat{H}_{average}(x_i) \parallel M) \biggr]
\end{aligned}
\end{equation}
In the end, we ascertain the total loss function for the module as follows:
\begin{equation}
\mathcal{L}_{prediction}=\mathcal{L}_{CrossEntropy}+\beta\mathcal{L}_{JS},
\end{equation}
the parameter $\beta$ is serves as a modulation coefficient to adjust the proportion of the JS loss.

\subsection{Cosine Classifier Loss}
\label{Cosine Classifier}

In classification tasks, class prototypes represent each class are determined by the mean embeddings. During the base session, prototypes are calculated using all samples in each class. For class $j$, and $K_{j}$ be the set of samples, the prototype $p_{j}$ is computed as $p_{j}=\frac{1}{|K_{j}|}\sum_{x_{i}\in K_{j}}f_{\theta}(x_{i})$. In incremental sessions, exemplars are used instead of all samples to calculate the prototypes. For class $j$ in session $t$, the prototype $p_{j}$ is is computed as $p_{j}=\frac{1}{|\mathcal{E}_{j}|}\sum_{x_{i}\in\mathcal{E}_{j}}f_{\theta}(x_{i})$, where $\mathcal{E}_{j}$ is the set of exemplars selected from \autoref{Exemplar Selection}.

Our novel loss formula integrates cosine similarity to emphasize directional alignment in the feature space and adaptability to evolving class distributions:
\begin{equation}
L_{CrossEntropy}\left(\theta ; x_i, y_i\right)=-\sum_{j=1}^{|C|} y_{i j} \log \left(\frac{\exp \left(\operatorname{sim}\left(f_\theta\left(x_i\right), p_j\right)\right)}{\sum_{k=1}^{|C|} \exp \left(\operatorname{sim}\left(f_\theta\left(x_i\right), p_k\right)\right)}\right),
\label{eq:CE}
\end{equation}
where $f_\theta(x_i)$ is the feature representation of $x_i$ extracted by the neural network, parameterized by $\theta$, and $p_j$ is the prototype of the $j$-th class. $sim(x,y)$ is defined in \autoref{sim}.

The overall loss function is defined as the sum of the SCL loss and the classification task loss, cross entropy loss:
\begin{equation}
\mathcal{L}=\mathcal{L}_{CrossEntropy}+\alpha\mathcal{L}_{SCL}
\end{equation}

\section{Experiments}
\subsection{Experimental Setup}
\subsubsection{Datasets}
MedMNIST \citep{medmnistv2} is a comprehensive and lightweight collection of biomedical image datasets aimed at facilitating the development and benchmarking of machine learning algorithms in biomedical imaging.
Our methodology was evaluated across two datasets from MedMNIST with their original resolutions: PathMNIST at $28\times28$ and BloodMNIST at $224\times224$. Statistical information of these two datasets are presented in \autoref{tab:dataset}. Following~\citet{openlongtailrecognition}, in imbalanced scenarios, sample sizes should range from 21 to 50, and in long-tailed scenarios, from 0 to 20.

\textbf{PathMNIST} \citep{kather2019predicting} is based on a dataset of histological images used for colorectal cancer research. It includes image patches of hematoxylin and eosin-stained tissue samples, encompassing nine distinct tissue types, making it suitable for multi-class classification tasks in histopathology.

\textbf{BloodMNIST} \citep{acevedo2020dataset} is composed of images of individual blood cells captured using a microscope, categorized into eight different classes, making it suitable for multi-class classification tasks and aiding in the development of automated systems for blood cell identification and analysis.

\begin{figure*}[!t]
\begin{center}
\includegraphics[width=0.9\textwidth]{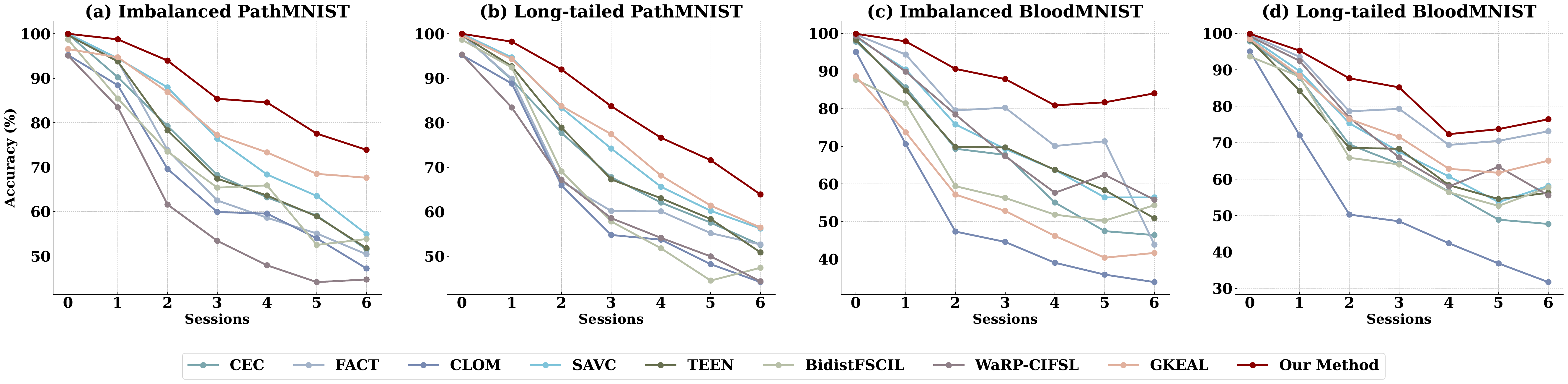}
 \caption{Comparison with SOTA methods on Imbalanced and Long-tailed PathMNIST, Imbalanced and Long-tailed BloodMNIST.}
 \label{fig:baseline}
 \end{center}
\end{figure*}

\subsubsection{Implementation Details}
We employed ResNet20 backbones for the $28\times28$ small-sized images in PathMNIST, while we utilize deeper ResNet18 backbones for the $224\times224$ larger-sized images in BloodMNIST. For both datasets, the SGD momentum coefficient was set to 0.9. For the PathMNIST dataset, the learning rate was set to 0.1 for the base session and 0.001 for the incremental sessions, with each session comprising 600 epochs. Conversely, for the BloodMNIST dataset, the learning rate was set to 0.002 for the base session and 0.000005 for the incremental sessions, with each session comprising 120 epochs. Specific choices regarding the hyperparameters for memory size and semantic expansion methods will be detailed in \autoref{Ablation Study}.

\subsubsection{Evaluation Metrics}
\textbf{1) Inter-class distance.} By maximizing the KL divergence between the probability distribution of new class samples and that of old class samples as the inter-class distance, we measure the separation between newly learned classes and previously learned classes \citep{li2023modeling}:
\begin{equation}
\label{equ:inter-class}
    L_{inter-class}=\frac12\left(D_{KL}(p_i||p_j)+D_{KL}(p_j||p_i)\right)
\end{equation}
Given that $p_i$ and $p_j$ represent the probability distributions of class $i$ and class $j$, respectively, the KL divergence $D_KL$ is defined as follows:
\begin{equation}
    D_{KL}(p_i\parallel p_j)=\sum_{k=1}^Cp_i(k)\log\frac{p_i(k)}{p_j(k)}
\end{equation}
\begin{equation}
    D_{KL}(p_j\parallel p_i)=\sum_{k=1}^Cp_j(k)\log\frac{p_j(k)}{p_i(k)}
\end{equation}

\textbf{2) Intra-class distance.} By minimizing the KL divergence between samples of the same class and their augmented counterparts as the intra-class distance, we evaluate the model's generalization ability to new classes. For class $i$:
\begin{equation}
\label{equ:intra-class}
    L_{intra\text{-}class}^i=\frac12\left(D_{KL}(p_i\parallel\hat{p}_i)+D_{KL}(\hat{p}_i\parallel p_i)\right)
\end{equation},
where $p_i$ and $\hat{p}_i$ represent the probability distributions of the original and augmented samples of class $i$, respectively.

\textbf{3) Model Uncertainty.} For each epoch, the entropies of all samples are averaged to obtain a measure of the overall model uncertainty:
\begin{equation}
\label{equ:model uncertainty}
    U=\frac1N\sum_{j=1}^NH(p^{(j)})
\end{equation}

\subsection{Comparison with SOTA methods}
Due to the lack of open-sourced class-incremental models specifically designed for limited sample sizes in the biomedical imaging domain, we applied the latest SOTA methods from the computer vision field to our datasets as baselines. These methods have been briefly described in \autoref{Related Work}, including CEC \citep{zhang2021few}, FACT \citep{zhou2022forward}, CLOM \citep{zou2022margin}, SAVC \citep{song2023learning}, TEEN \citep{wang2023fewshot}, BiDistFSCIL \citep{zhao2023few}, WaRP-CIFSL \citep{kim2023warping} and GKEAL \citep{Zhuang_GKEAL_CVPR2023}. Detailed experimental data for the imbalanced BloodMNIST dataset are presented in \autoref{tab:sota}, while comparative results for other datasets are shown in \autoref{fig:baseline}.

\begin{table}[!t]
\scriptsize
\setlength\tabcolsep{3pt}
  \centering
  \caption{Comparison with SOTA methods on imbalanced BloodMNIST dataset. $\Delta_{\text{final}}$ represents the value by which our method surpasses the accuracy of methods in the baseline during the final session. $\Delta_{\text{average}}$ represents the value by which our method exceeds the average accuracy of the methods in the baseline.}
  \label{tab:sota}
\renewcommand{\arraystretch}{1.25}
    \begin{tabular}{ccccccccccccc}
    \toprule

    \multirow{2}{*}{Method} & \multicolumn{7}{c}{Accuracy in each session (\%) ↑}                                      & \multirow{2}{*}{$\Delta_{\text{final}}$} & \multirow{2}{*}{$\Delta_{\text{average}}$}\\
\cline{2-8}          & 0     & 1     & 2     & 3     & 4     & 5     & 6     &      &      &      \\
    \hline
    CEC & 97.78  & 85.66  & 69.34  & 67.71  & 55.04  & 47.45  & 46.39  & -37.67 & -21.92 \\
    FACT & 99.67  & 94.37  & 79.54  & 80.24  & 70.07  & 71.31  & 73.85  & -10.21 & -7.53 \\
    CLOM  & 95.05  & 70.57  & 47.33  & 44.58  & 39.04  & 35.89  & 30.52  & -53.54 & -37.12 \\
    SAVC & 98.95  & 90.39  & 75.80  & 69.29  & 63.73  & 56.41  & 60.42  & -23.64 & -15.40 \\
    TEEN & 98.28  & 84.80  & 69.77  & 69.62  & 58.63  & 54.74  & 56.17  & -27.89 & -18.68 \\
    BidistFSCIL  & 87.67  & 81.42  & 59.39  & 56.27  & 51.82  & 50.22  & 54.34  & -29.72 & -25.95 \\
    WaRP-CIFSL & 99.31 & 89.82 & 78.44 & 67.37 & 57.64 & 62.42 & 55.80 & -28.26& -16.00 \\
    GKEAL   & 88.60  & 73.71  & 57.17  & 52.82  & 46.17  & 40.39  & 41.65  & -42.41 & -31.75 \\
    \midrule
    \textbf{Our Method}  & \textbf{99.89}  & \textbf{97.87}  & \textbf{90.56}  & \textbf{87.86}  & \textbf{80.86}  & \textbf{81.68}  & \textbf{84.06}  & \textbf{$-$} & \textbf{88.97} \\
    \bottomrule
    \end{tabular}%
\end{table}%

From~\autoref{tab:sota} and~\autoref{fig:baseline}, our method significantly outperforms other SOTA methods from the computer vision field over the past three years in terms of average accuracy and accuracy in the last session, particularly under conditions of long-tail and imbalanced data distributions. This superiority is due to the distinct differences between biomedical and natural images and confirms that models pre-trained on natural images can not directly applied to biomedical image classification tasks.

\begin{table*}[htbp]
\scriptsize
  \centering
  \caption{Ablation studies on long-tailed BloodMNIST dataset. \textbf{F}, \textbf{E}, \textbf{C} denote Fine-Grained Semantic Expansion, Method for Selecting Exemplars, and Classifiers, respectively. And ``UTA'' refers to selecting samples using Uncertainty Trajectory Analyzer; ``RANDOM'' denotes random selection; ``NME'' stands for Nearest-Mean-of-Exemplars strategy; ``POOL'' and ``COMMITTEE'' are pool-based and committee-based sampling, respectively. $\Delta_{\text{final}}$ represents to the comparison of the decline in accuracy of our method in the last session. `average' indicates the mean accuracy across all sessions. $\Delta_{\text{average}}$ denotes the comparison of the decline in the average accuracy of our method.}
  \label{tab:ablation}

\renewcommand{\arraystretch}{1.25}
    \begin{tabular}{cccccccccccccc}
    \toprule
    \multirow{2}*{\textbf{}} & \multirow{2}*{\textbf{F}} & \multirow{2}*{\textbf{E}} & \multirow{2}*{\textbf{C}} & \multicolumn{7}{c}{Accuracy in each session (\%) ↑}                       & \multirow{2}{*}{$\Delta_{\text{final}}$} & \multirow{2}{*}{average} & \multirow{2}{*}{$\Delta_{\text{average}}$}\\
\cline{5-11}          &       &       &       & 0     & 1     & 2     & 3     & 4     & 5     & 6     &      &  \\
    \hline
    Ours      &\checkmark       &UTA       &cos       & \textbf{99.89}  & \textbf{95.26}  & \textbf{87.68}  & \textbf{85.19}  & \textbf{72.34}  & \textbf{73.74}  & \textbf{76.43}  & $-$  & \textbf{84.36}  & $-$\\
    \hdashline
    w/o F     &       &UTA       &cos       & 98.78  & 89.04  & 74.62  & 68.81  & 58.42  & 56.25  & 58.17  & -18.16  & 72.01  & -12.35\\
    \hdashline
    w/o E      & \checkmark     &RANDOM      &cos       & 98.89  & 88.06  & 72.95  & 69.33  & 52.09  & 52.83  & 57.62  & -18.18  & 70.25  & -14.11\\
         & \checkmark     &NME     &cos       & 98.39  & 90.40  & 71.80  & 66.48  & 60.93  & 53.00  & 58.72  & -17.71  & 71.39  & -12.97\\
         & \checkmark     &POOL     &cos       & 99.56  & 92.08  & 75.71  & 72.91  & 62.83  & 56.57  & 61.35  & -15.08  & 74.43  & -9.93\\
         & \checkmark     &COMMITTEE     &cos       & 99.67  & 92.50  & 76.31  & 75.00  & 68.06  & 65.62  & 65.66  & -10.77  & 77.55  & -6.81\\
    \hdashline
    w/o C      & \checkmark     &UTA      &dot       & 99.89  & 49.17  & 34.96  & 23.81  & 29.88  & 32.21  & 37.76  & -38.67  & 43.95  & -40.41\\
         & \checkmark     &UTA     &euc       & 98.23  & 87.05  & 65.34  & 70.67  & 57.21  & 54.40  & 57.31  & -19.12  & 70.03  & -14.33\\
         & \checkmark     &UTA     &mah       & 100.0  & 82.06  & 56.16  & 52.29  & 36.17  & 36.04  & 46.23  & -30.2  & 58.42  & -25.94\\
    \bottomrule
    \end{tabular}%
  \label{tab:ablate}%
\end{table*}%


\subsection{Ablation Study}\label{Ablation Study}
We conduct ablation studies to validate the efficacy of the three modules we proposed: Fine-Grained Semantic Expansion in~\autoref{Denser Semantic Space}, Uncertainty Trajectory Analyzer in~\autoref{Exemplar Selection}, and the Cosine Classifier in~\autoref{Cosine Classifier}. Inspired by active learning principles, we also compare our approach with traditional pool-based~\citep{settles2009active} and committee-based~\citep{settles2009active} active learning methods. In terms of classifiers, we evaluate the performance differences between cosine similarity and other metrics such as dot product, Euclidean distance~\citep{wang2005euclidean}, and Mahalanobis distance~\citep{haasdonk2010classification}. The results of the ablation studies conducted on the long-tailed BloodMNIST dataset are presented in~\autoref{tab:ablation}.

\begin{figure}[!t]
\centering
\includegraphics[width=0.5\textwidth]{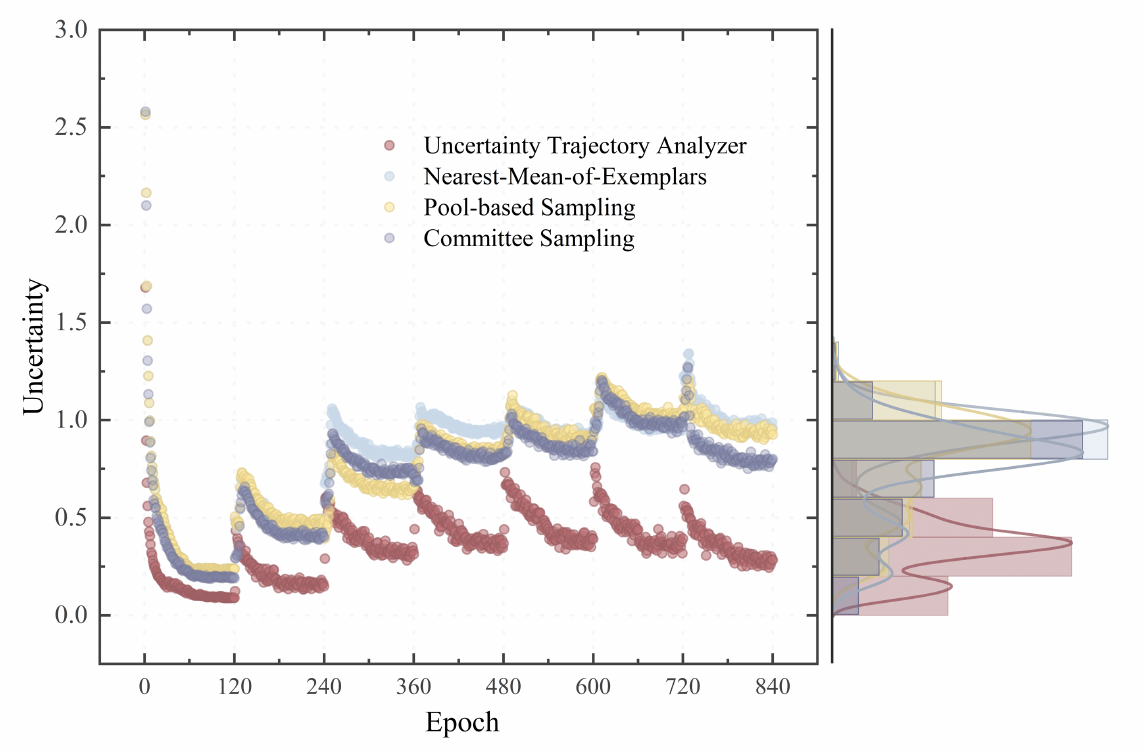}
\caption{On the imbalanced BloodMNIST dataset, different exemplar selection strategies influence the overall distribution of model uncertainty across all training cycles. The density estimation plot on the right provides additional details about the distribution of uncertainty. The initial 120 epochs constitute the base session, followed by six incremental sessions, each also comprising 120 epochs.}
\label{fig:uncertainty_epoch}
\end{figure}

Results indicate that our Fine-Grained Semantic Expansion enhances the model generalization capability, it exceeds the control group by 18.16\% and 12.35\% in the last session as well as in the average accuracy, respectively. Concerning exemplar selection, NME provides stable performance in some sessions but may be inadequate for handling highly uneven intra-class data distributions. Pool-based active learning strategy selects the most informative samples from a predefined pool for labeling, with its performance influenced by the accuracy of heuristic evaluations, showing less stability in some sessions compared to UTA. Our UTA outperforms RANDOM by 14.11\%, NME by 12.97\%, POOL by 9.93\%, COMMITTEE by 6.81\% in terms of average accuracy. Regarding classifier performance, cosine similarity exhibits superior discriminative ability when dealing with unevenly distributed features in imbalanced and long-tailed datasets.

\subsection{Further Analyses via Visualization}

\textbf{\Rmnum{1}. Confusion Matrix.}

\begin{figure*}[!t]
\begin{center}
\includegraphics[width=1\textwidth]{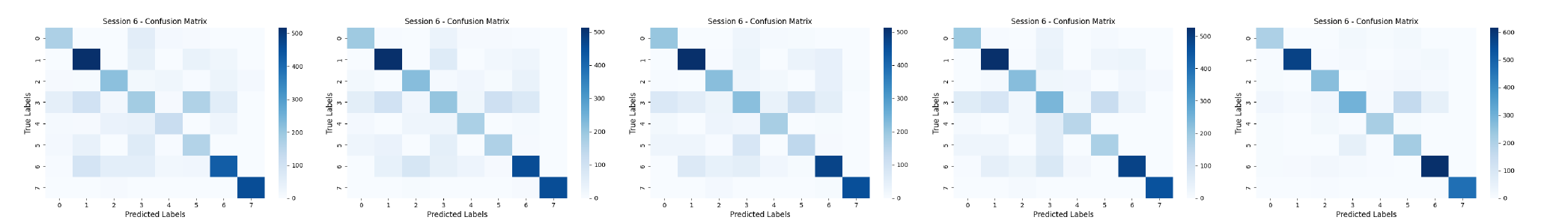}
\begin{tikzpicture}[overlay]
    \node at (-7.3,0.2) {\scriptsize (a) Random};
    \node at (-3.9,0.2) {\scriptsize (b) NME};
    \node at (-0.3,0.2) {\scriptsize (c) Pool-based};
    \node at (3.2,0.2) {\scriptsize (d) Committee};
    \node at (6.7,0.2) {\scriptsize (e) UTA};
\end{tikzpicture}
 \caption{Confusion matrices for different exemplar selection methods on the imbalanced BloodMNIST dataset.}
 \label{fig:matrix}
 \end{center}
\end{figure*}
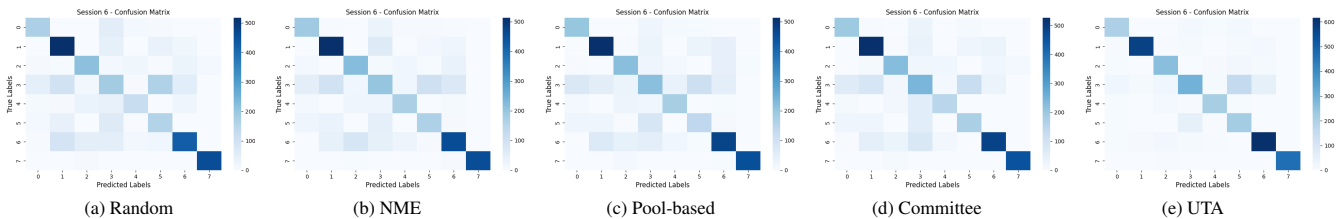

We visualized the confusion matrices from the final session on the imbalanced BloodMNIST dataset, comparing different exemplar selection methods. Results are illustrated in \autoref{fig:matrix}.

The shading in the confusion matrices illustrates the degree of match between predicted and true labels. Random selection shows dispersed results. Despite NME selection has high accuracy in certain classes, it falls short in recognizing smaller ones. Moreover, Pool-based and Committee-based strategies exhibit good matches in some classes but misclassifications in others suggest potential inconsistencies in the models. UTA is highly concentrated in the confusion matrix, surpassing other methods.

\textbf{\Rmnum{2}. Model Uncertainty.}

\autoref{fig:uncertainty_epoch} illustrates the impact of different exemplar selection strategies on the model uncertainty across the entire training cycle on the imbalanced BloodMNIST dataset. For each epoch, averaging the entropy across all samples yields a measure of the model's overall uncertainty in~\autoref{equ:model uncertainty}.

A rapid decline in model uncertainty indicates that the model is quickly approaching a state of convergence, hence learning more efficiently. Furthermore, as each session progresses with increasing epochs, the model's uncertainty decreases, and, coupled with rising prediction accuracy as shown in~\autoref{tab:ablation}, the model's predictive power strengthens. Compared to other curves and distributions, our UTA tends to focus more on areas with lower uncertainty, yielding more certain and accurate predictions, and demonstrates a steeper decline, indicating stronger learning and predictive capabilities.

\textbf{\Rmnum{3}. Class Separation Diagram.}

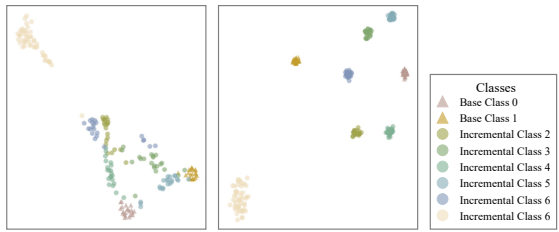
\begin{figure}[!t]
\centering
\includegraphics[width=0.4\textwidth]{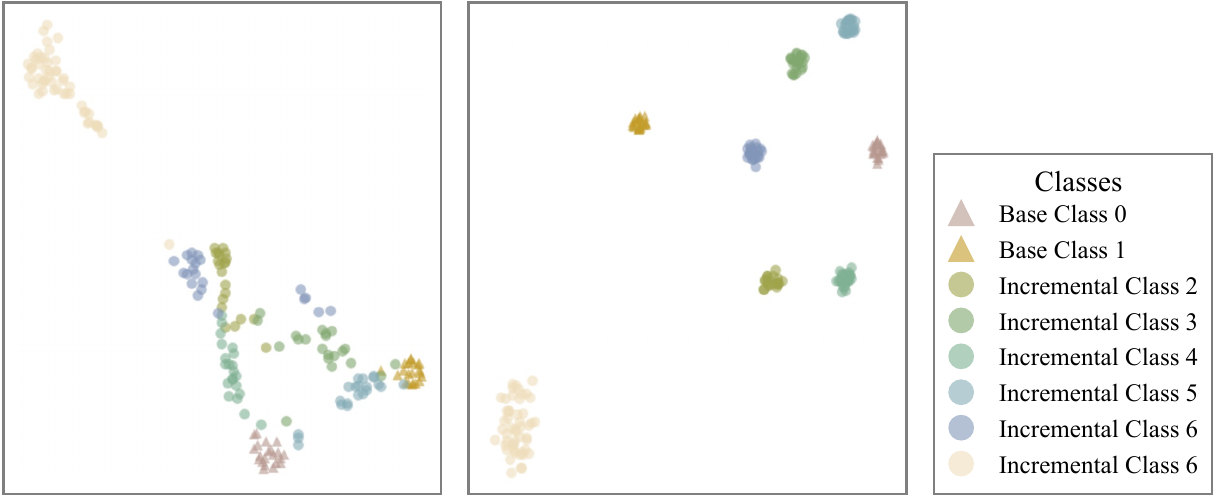}
\begin{tikzpicture}[overlay]
    \node at (-6.3,3.3) {\scriptsize (a) Without Semantic Expansion};
    \node at (-3.2,3.3) {\scriptsize (b) With Semantic Expansion};
\end{tikzpicture}
\caption{t-SNE visualization for the imbalanced BloodMNIST datasets under conditions with and without Fine-Grained Semantic Expansion.}
\label{fig:tsne}
\end{figure}

We visualize the feature spaces of the final session for the imbalanced PathMNIST and BloodMNIST datasets using t-SNE, comparing the scenarios with and without Fine-Grained Semantic Expansion. The results are shown in \autoref{fig:tsne}.

Without Fine-Grained Semantic Expansion, feature points are more dispersed and the separation between classes is less distinct, resulting in unclear boundaries between classes. Conversely, with its use, overlap between classes is reduced, and issues of inter-class confusion are mitigated, enhancing both intra-class cohesion and inter-class separability and providing a richer semantic space.

\textbf{\Rmnum{4}. Class Distribution Network Diagram.}

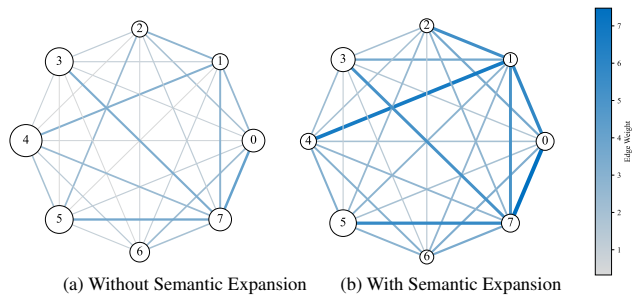
\begin{figure}[!t]
\centering
\includegraphics[width=0.45\textwidth]{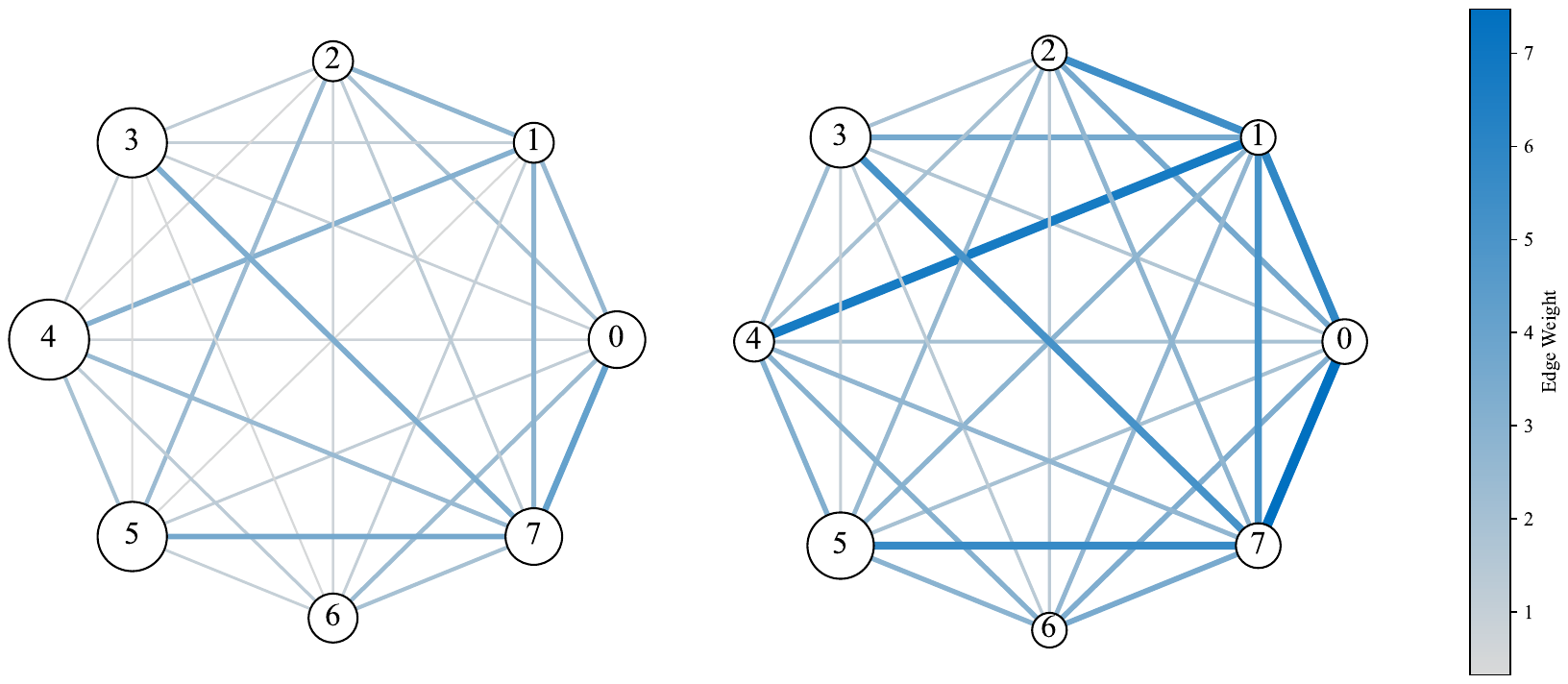}
\begin{tikzpicture}[overlay]
    \node at (-6,-0.1) {\scriptsize (a) Without Semantic Expansion};
    \node at (-2.5,-0.1) {\scriptsize (b) With Semantic Expansion};
\end{tikzpicture}
\caption{Network graph of classes on the imbalanced BloodMNIST dataset with and without Fine-Grained Semantic Expansion: Each node represents a class, where the size of the node's circle indicates the compactness within the category—the smaller the circle, the shorter the intra-class distance. The thickness and color intensity of the lines between nodes denote the level of dispersion between classes, with thicker and darker lines indicating greater distances between classes.}
\label{fig:network}
\end{figure}

From a more detailed perspective, on the imbalanced BloodMNIST dataset, we utilize~\autoref{equ:inter-class} to measure inter-class distances between different classes and intra-class distances between original samples and their virtually augmented counterparts within the same class by~\autoref{equ:intra-class}. This analysis facilitates the construction of a network graph between classes, serving as a supplement to the t-SNE visualization of the feature space. We continue to compare scenarios with and without Fine-Grained Semantic Expansion, with results shown in~\autoref{fig:network}. In the graph, each node represents a class; the size of a node’s circle indicates the compactness within the class—the smaller the circle, the smaller the intra-class distance. The thickness and color depth of the lines between nodes signify the degree of dispersion between classes, with thicker and darker lines indicating greater distances between classes. This visualization further confirms that after introducing Fine-Grained Semantic Expansion, the feature embeddings of all classes are more compactly populated by virtually augmented samples, leaving space for potential generalization to new classes.

\textbf{\Rmnum{5}. Mitigating Class Imbalance with a Cosine Classifier.}

\begin{figure}[!t]
\centering
\includegraphics[width=0.5\textwidth]{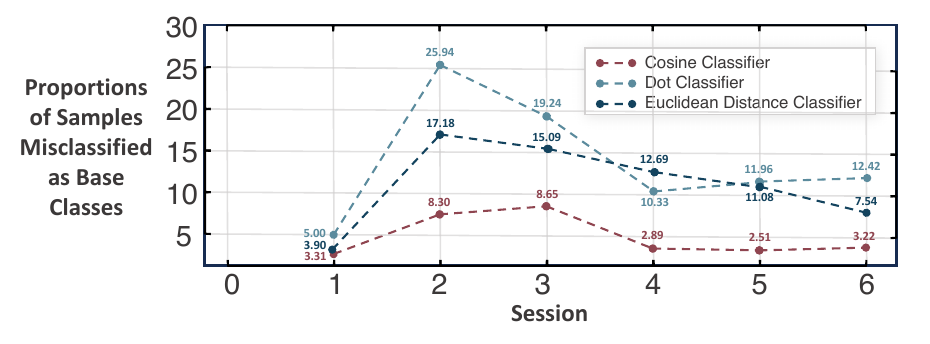}
\caption{The proportion of samples from newly introduced classes in the long-tailed BloodMNIST dataset that are misclassified as base classes when using different classifiers.}
\label{fig:classifier}
\end{figure}

When utilizing various classifiers on long-tailed BloodMNIST dataset, the proportion of samples from newly introduced classes that are misclassified as base classes changes over the course of training visualized in~\autoref{fig:classifier}. Compared to other classifiers, Cosine Classifier effectively prevents the shift of weights towards previous classes and significantly mitigates the issue of new samples being misclassified as base classes.

\subsection{More Analyses for Hyper-parameters}
\subsubsection{Semantic Expansion for Biomedical Image}
\begin{figure}[!t]
\centering
\includegraphics[width=0.5\textwidth]{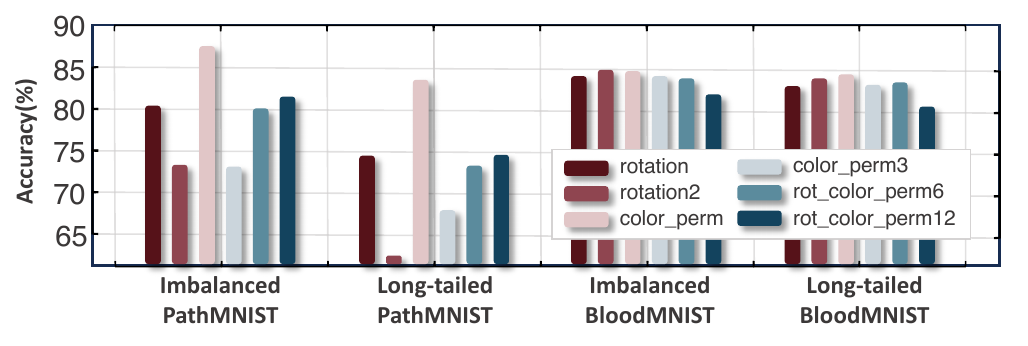}
\caption{Classification accuracy using different semantic expansion methods.}
\label{fig:fantasy}
\end{figure}

Research by~\citet{zhang2023unified} and \citet{zhang2024imperceptiblebackdoorattackselfsupervised} indicated that ColorJitter could impair the detailed textural features of biomedical images. Based on experimental results in \autoref{fig:fantasy}, we chose different semantic expansion methods for various datasets: ``color\_perm'' transformation for both the imbalanced and long-tailed PathMNIST and long-tailed BloodMNIST datasets, and the ``rotation2'' transformation for the imbalanced BloodMNIST dataset. 

``rotation'' applies rotations at four angles (0, 90, 180, 270 degrees), while ``rotation2'' restricts this to two (0, 180 degrees); ``color\_perm'' outputs all possible color channel permutations, whereas ``color\_perm3'' only generates two additional versions; ``rot\_color\_perm6'' combines two rotations (0, 180 degrees) with three color permutations, and ``rot\_color\_perm12'' expands this by including all four rotational degrees each paired with three color permutations.

\subsubsection{Memory Size}
\begin{figure}[!t]
\centering
\includegraphics[width=0.5\textwidth]{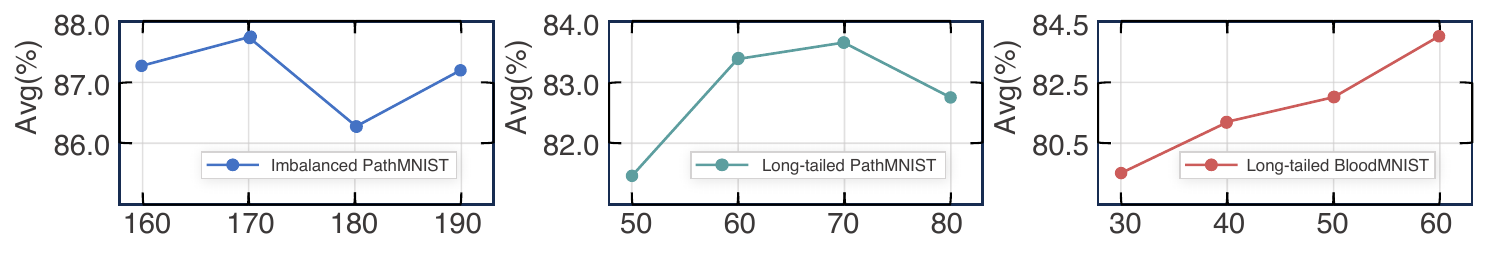}
\caption{The impact of varying memory sizes on classification accuracy across different datasets.}
\label{fig:memory}
\end{figure}

Excessively large memory sizes can result in the allocation of more samples per class than available in the dataset, while excessively small sizes may include too few samples for effective model training. Therefore, the memory size for different datasets is initially constrained within a specific range. Subsequently, optimal memory sizes are experimentally determined for each dataset. As shown in \autoref{fig:memory}, for the imbalanced PathMNIST, we select a memory size of m=170; for the long-tailed PathMNIST, m=70; for the imbalanced BloodMNIST, m=150; and for the long-tailed BloodMNIST, m=60.

\section{Conclusion}
In this paper, we propose a novel method dubbed ESSENTIAL, focusing on scenarios involving imbalanced and long-tailed data. Our UTA module first validates the effectiveness of cumulative entropy in measuring sample uncertainty and leverages this concept for predicting the most uncertain samples to be selected as exemplars. Besides, our Fine-Grained Semantic Expansion module enhances the supervised contrastive learning model by incorporating various semantic expansions at a finer granularity, enriching the semantic space and improving generalization to new classes. Finally, the cosine classifier further enhances the model's ability to handle class imbalance. Our method surpasses state-of-the-art methods and becomes the first class-incremental learning method under limited samples in the field of biomedical imaging.


\bibliographystyle{model2-names.bst}\biboptions{authoryear}
\bibliography{refs}

\end{document}